\documentclass[journal,twocolumn]{IEEEtran}

\usepackage{amsmath,amssymb,amsfonts}
\usepackage{graphicx}
\usepackage{cite}
\usepackage{booktabs} 
\usepackage[hyphens]{url}
\usepackage{hyperref}
\usepackage{float}    
\usepackage{stfloats} 

\begin{document}

\title{L-JacobiNet and S-JacobiNet: An Analysis of Adaptive Generalization, Stabilization, and Spectral Domain Trade-offs in GNNs}

\author{
    Hüseyin Göksu,~\IEEEmembership{Member,~IEEE}
    \thanks{H.
Göksu, Akdeniz Üniversitesi, Elektrik-Elektronik Mühendisliği Bölümü, Antalya, Türkiye, e-posta: hgoksu@akdeniz.edu.tr.}%
    \thanks{Manuscript received October 31, 2025;
revised XX, 2025.}
}

\markboth{IEEE TRANSACTIONS ON SIGNAL PROCESSING, VOL. XX, NO.
XX, NOVEMBER 2025}
{Göksu: L-JacobiNet and S-JacobiNet: An Analysis}

\maketitle

\begin{abstract}
Spectral GNNs, like ChebyNet, are limited by heterophily and over-smoothing due to their static, low-pass filter design.
This work investigates the "Adaptive Orthogonal Polynomial Filter" (AOPF) class as a solution.
We introduce two models operating in the [-1, 1] domain: 1) `L-JacobiNet`, the adaptive generalization of `ChebyNet` with learnable $\alpha, \beta$ shape parameters, and 2) `S-JacobiNet`, a novel baseline representing a LayerNorm-stabilized static `ChebyNet`.
Our analysis, comparing these models against AOPFs in the [0, $\infty$) domain (e.g., `LaguerreNet`), reveals critical, previously unknown trade-offs.
We find that the $[0, \infty)$ domain is superior for modeling heterophily, while the [-1, 1] domain (Jacobi) provides superior numerical stability at high K (K>20).
Most significantly, we discover that `ChebyNet`'s main flaw is \textit{stabilization}, not its \textit{static} nature.
Our static `S-JacobiNet` (ChebyNet+LayerNorm) outperforms the adaptive `L-JacobiNet` on 4 out of 5 benchmark datasets, identifying `S-JacobiNet` as a powerful, overlooked baseline and suggesting that adaptation in the [-1, 1] domain can lead to overfitting.
\end{abstract}


\section{INTRODUCTION}

\textbf{S}pectral Graph Neural Networks (GNNs), rooted in Graph Signal Processing (GSP) \cite{shuman2013emerging}, define graph convolutions as filters operating on the graph Laplacian spectrum.
The foundational model, `ChebyNet` \cite{defferrard2016convolutional}, approximates a filter $g_\theta(L)$ with a truncated expansion of Chebyshev polynomials $P_k(L)$.
This static, low-pass design leads to two fundamental problems:
\begin{enumerate}
    \item \textbf{Failure on Heterophily:} The low-pass filter fails on heterophilic graphs, where high-frequency signals (dissimilar neighbors) dominate \cite{HeroFilter2025, zhu2020beyond}.
\item \textbf{Over-smoothing:} The filter's low-pass nature intensifies with $K$, causing performance to collapse at high degrees \cite{li2018deeper}.
\end{enumerate}

To solve this, we recently proposed a class of \textbf{Adaptive Orthogonal Polynomial Filters (AOPF)} \cite{goksu2025meixnernet, goksu2025krawtchouknet, goksu2025laguerrnet}, which learn the filter's shape parameters ($\alpha, \beta, p$, etc.).
Our prior work focused on the $[0, \infty)$ domain (e.g., `MeixnerNet`, `LaguerreNet`).
In this work, we conduct a foundational analysis of the AOPF framework by focusing on the $[-1, 1]$ domain, the home of `ChebyNet`.
We introduce and analyze two models:
\begin{enumerate}
    \item \textbf{`L-JacobiNet`:} The adaptive generalization of `ChebyNet`.
It uses Jacobi polynomials $P_k^{(\alpha, \beta)}(x)$ and makes the $\alpha, \beta$ shape parameters learnable.
    \item \textbf{`S-JacobiNet`:} Our novel ablation baseline.
It is a `ChebyNet` filter (static $\alpha=\beta=-0.5$) stabilized using the `LayerNorm` framework from our AOPF class.
\end{enumerate}

Our extensive analysis (Section IV) of these two models against the AOPF class yields three critical, non-trivial findings:

\begin{enumerate}
    \item \textbf{Domain Trade-off (Heterophily):} The $[0, \infty)$ domain AOPFs (`LaguerreNet`, `MeixnerNet`) are superior for modeling heterophily (Table II).
\item \textbf{Domain Trade-off (Stability):} The $[-1, 1]$ domain (`L-JacobiNet`) provides superior numerical stability at high K ($K=30$), whereas the $[0, \infty)$ `LaguerreNet` collapses.
\item \textbf{Adaptation vs. Stabilization Trade-off:} Our most significant finding. The static `S-JacobiNet` (ChebyNet+LayerNorm) outperforms the adaptive `L-JacobiNet` on 4 out of 5 datasets (Table IV).
This suggests `ChebyNet`'s main flaw was \textit{stabilization}, not its \textit{static} nature, and that adaptivity in the $[-1, 1]$ domain may lead to overfitting.
\end{enumerate}

This paper shifts the GNN narrative from "finding one best filter" to "understanding the crucial trade-offs" between spectral domain, adaptation, and stabilization.
\section{RELATED WORK}
Our work intersects three research areas: spectral filter design, solutions for heterophily, and solutions for over-smoothing.
\subsection{Spectral Filter Design in GNNs}
Spectral GNN filters $g_\theta(L)$ fall into several classes:
\begin{itemize}
    \item \textbf{Static Polynomial (FIR) Filters:} The most common class, including `ChebyNet` \cite{defferrard2016convolutional} (Chebyshev), `GCN` \cite{kipf2017semi}, and `BernNet` \cite{he2021bernnet} (Bernstein).
\item \textbf{Static Basis + Learned Coefficients:} This class fixes the basis but learns the $\theta_k$ coefficients.
`APPNP` \cite{gasteiger2019predict} and `GPR-GNN` \cite{chien2021adaptive} are prime examples. The static `JacobiConv` (Wang et al., 2022) \cite{JacobiConv2022} also fits this class, using a static Jacobi basis for its flexibility.
\item \textbf{Rational (IIR) Filters:} More complex filters using ratios of polynomials, such as `CayleyNet` \cite{levie2018cayleynets} and `ARMAConv` \cite{ARMAConv2021}.
\end{itemize}
\textbf{Our Approach: Adaptive Basis Filters (AOPF).} `L-JacobiNet` belongs to a fourth class we introduced \cite{goksu2025meixnernet, goksu2025krawtchouknet, goksu2025laguerrnet}.
We do not learn $\theta_k$; we learn the polynomial's fundamental shape parameters ($\alpha, \beta$).
Our work directly contrasts with the static `JacobiConv` \cite{JacobiConv2022} by 1) making the basis itself learnable (`L-JacobiNet`) and 2) introducing robust `LayerNorm` stabilization (`S-JacobiNet`).
\subsection{Solutions for Heterophily}
Heterophily requires capturing high-frequency signals. Solutions include architectural changes (`H2GCN` \cite{zhu2020beyond}) or adaptive aggregation (`GAT` \cite{velickovic2018graph}).
Recent work explicitly links heterophily to the need for "band-pass" or "high-pass" filters, a hypothesis our work confirms (Section IV.B).
\subsection{Solutions for Over-smoothing and Stabilization}
Over-smoothing (high-$K$ collapse) is often solved architecturally (`GCNII` \cite{chen2020simple}). Filter-level stabilization is less explored.
`ChebyNet` stabilization has focused on residual connections (e.g., `ResChebyNet` \cite{ResChebyNet2024}) or Lipschitz normalization.
Our work demonstrates that a simple `LayerNorm` \cite{ba2016layer} application, in contrast to these architectural solutions, provides potent and direct \textit{filter level} stabilization.
The success of our `S-JacobiNet` suggests this simple but powerful baseline has been overlooked by the GNN community.

THE AOPF FRAMEWORK AND DOMAIN ANALYSIS
\section{The AOPF Framework and Domain Analysis}
We define our AOPF models based on their polynomial basis and spectral domain.
All models use the same architecture (Section IV.A) and `LayerNorm` stabilization.
\subsection{Domain 1: The $[0, \infty)$ (Semi-Infinite) Domain}
These filters map the Laplacian $L_{sym}$ to the $[0, 1]$ range via $L_{scaled} = 0.5 \cdot L_{sym}$.
\begin{itemize}
    \item \textbf{`LaguerreNet` \cite{goksu2025laguerrnet}:} (Continuous) Uses Laguerre polynomials $L_k^{(\alpha)}(x)$ with a learnable $\alpha$.
Coefficients $c_k \sim O(k^2)$ (unbounded).
    \item \textbf{`MeixnerNet` \cite{goksu2025meixnernet}:} (Discrete) Uses Meixner polynomials $M_k(x; \beta, c)$ with learnable $\beta, c$.
Also $O(k^2)$ unbounded.
    \item \textbf{`KrawtchoukNet` \cite{goksu2025krawtchouknet}:} (Discrete) Uses Krawtchouk polynomials $K_k(x; p, N)$ with learnable $p$.
$N$ is fixed, making coefficients \textit{bounded}.
\end{itemize}

\subsection{Domain 2: The $[-1, 1]$ (Finite) Domain}
These filters map the Laplacian $L_{sym}$ to the $[-1, 1]$ range via $L_{hat} = L_{sym} - I$ (assuming $\lambda_{max}=2$).
\begin{itemize}
    \item \textbf{`ChebyNet` \cite{defferrard2016convolutional}:} (Static, Unstable) Uses Chebyshev polynomials (Jacobi with $\alpha=\beta=-0.5$) and lacks stabilization.
\item \textbf{`L-JacobiNet` (This work):} (Adaptive, Stable) Uses Jacobi polynomials $P_k^{(\alpha, \beta)}(x)$ with learnable $\alpha > -1$ and $\beta > -1$ and `LayerNorm` stabilization.
Its $O(k^2)$ coefficients are \textit{unbounded}.
    
    \item \textbf{`S-JacobiNet` (This work):} (Static, Stable) Our ablation model.
It is `L-JacobiNet` with $\alpha$ and $\beta$ permanently fixed to $-0.5$.
It is functionally a `ChebyNet` filter combined with our `LayerNorm` stabilization framework.
\end{itemize}

EXPERIMENTAL ANALYSIS
\section{EXPERIMENTAL ANALYSIS}
We now present the experimental results from the v3 Colab run, structured around the three trade-offs we discovered.
\subsection{Experimental Setup}
\textbf{Datasets:} We use homophilic (Cora, CiteSeer, PubMed) and heterophilic (Texas, Cornell) benchmarks.
\textbf{Baselines:} We test our new models (`L-JacobiNet`, `S-JacobiNet`) against the AOPF class (`MeixnerNet`, `KrawtchoukNet`, `LaguerreNet`) and SOTA (`ChebyNet`, `GAT`, `APPNP`).
\textbf{Training:} All models use a 2-layer `PolyBaseModel` structure with $H=16$ (unless noted) and $K=3$ (for heterophily/homophily) or $K$ up to 30 (for over-smoothing).
\begin{figure*}[t]
\centerline{\includegraphics[width=\textwidth, height=0.85\textheight, keepaspectratio]{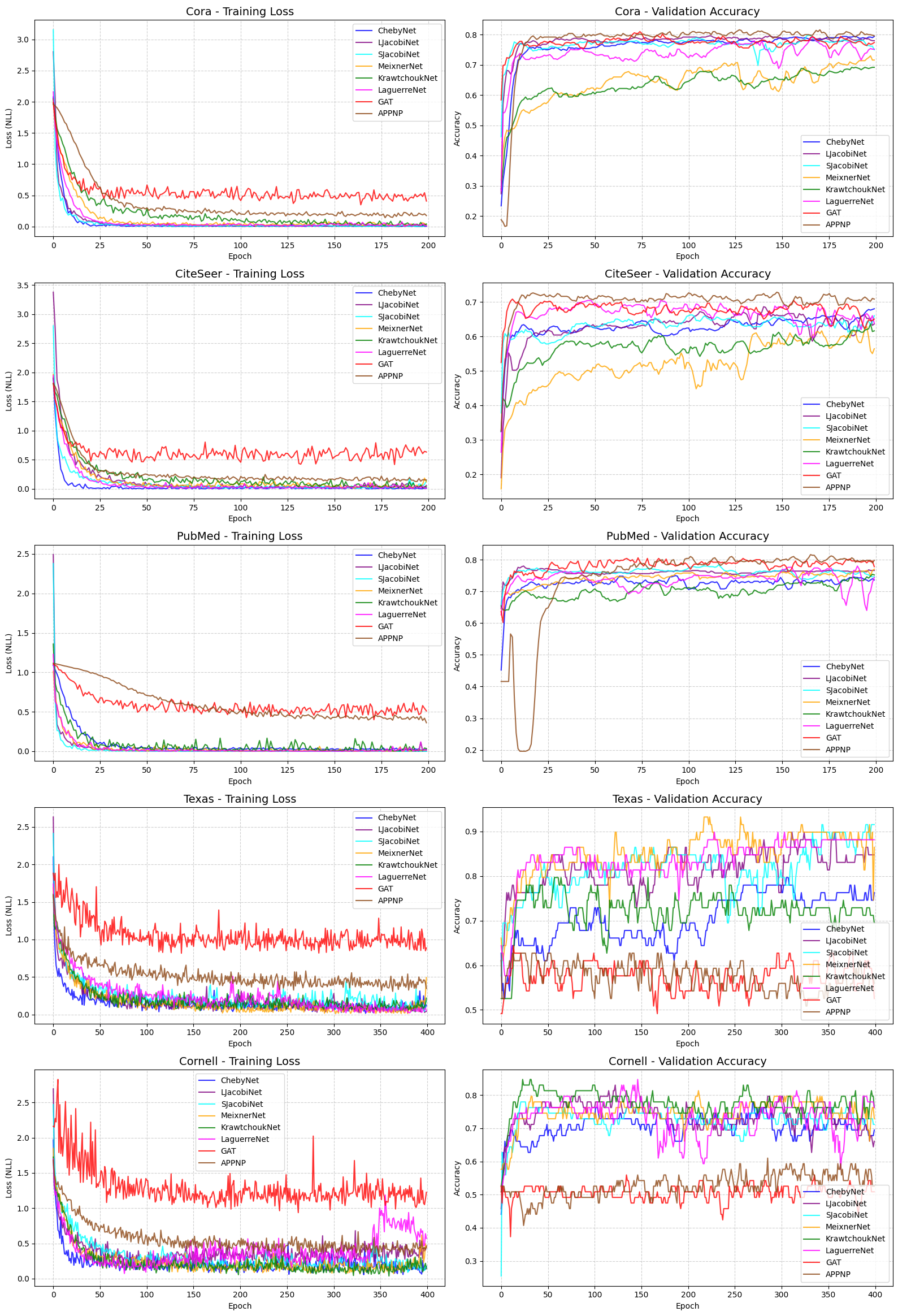}}
\caption{Figure 1: Training dynamics comparison (K=3, H=16).
On heterophilic datasets (Texas, Cornell), the $[-1, 1]$ domain filters (`ChebyNet`, `LJacobiNet`, `S-JacobiNet`) and standard baselines (`GAT`, `APPNP`) struggle, while the $[0, \infty)$ domain filters (`MeixnerNet`, `KrawtchoukNet`, `LaguerreNet`) are visibly more stable and accurate.}
\label{fig:training_curves}
\end{figure*}

\subsection{Trade-off 1: Heterophily vs. Spectral Domain}
We first analyze the filter's ability to handle heterophily.
Table \ref{tab:homophilic_results} shows performance on standard homophilic benchmarks, where `APPNP` excels. Table \ref{tab:heterophilic_results} shows the results for heterophily.
\begin{table*}[t]
\caption{Test accuracies (\%) on homophilic datasets (K=3, H=16).}
\label{tab:homophilic_results}
\centering
\begin{tabular}{l c c c c c c c c}
\toprule
\textbf{Model} & \textbf{ChebyNet} & \textbf{LJacobiNet} & \textbf{SJacobiNet} & \textbf{MeixnerNet} & \textbf{KrawtchoukNet} & \textbf{LaguerreNet} & \textbf{GAT} & \textbf{APPNP} \\
\midrule
Cora & 0.7990 & 0.7840 & 0.7840 & 0.7220 & 0.6950 & 0.7900 & 0.8220 & \textbf{0.8380} \\
CiteSeer & 0.6720 & 0.6580 & 0.6600 & 0.6110 & 0.6350 & 0.6830 & 0.6840 & \textbf{0.7110} \\
PubMed & 0.7320 & 0.7550 & 0.7520 & 0.7730 & 0.7330 & 0.7730 & 0.7710 & \textbf{0.7880} \\
\bottomrule
\end{tabular}
\end{table*}

SÜTUNA DÜZELTİLDİ) --------
\begin{table*}[t]
\caption{Test accuracies (\%) on heterophilic datasets (K=3, H=16).
10-fold Mean.}
\label{tab:heterophilic_results}
\centering
\begin{tabular}{l c c c c c c c c}
\toprule
\textbf{Model} & \textbf{ChebyNet} & \textbf{LJacobiNet} & \textbf{SJacobiNet} & \textbf{MeixnerNet} & \textbf{KrawtchoukNet} & \textbf{LaguerreNet} & \textbf{GAT} & \textbf{APPNP} \\
\midrule
Texas & 0.7135 & 0.8000 & 0.7811 & \textbf{0.8730} & 0.7432 & 0.8297 & 0.5919 & 0.5757 \\
Cornell & 0.6432 & 0.6378 & 0.6757 & \textbf{0.7162} & 0.6919 & 0.6730 & 0.4676 & 0.4459 \\
\bottomrule
\end{tabular}
\end{table*}

\textbf{Analysis of Heterophily Results:}
The results in Table \ref{tab:heterophilic_results} reveal a clear domain-specific trade-off.
\begin{itemize}
    \item Standard baselines (`GAT`, `APPNP`) completely fail, as their low-pass bias is fundamentally mismatched with the high-frequency signals of heterophily.
This is visually confirmed in Figure \ref{fig:training_curves} (bottom rows), where their validation accuracy is low and erratic.
\item The $[-1, 1]$ domain filters (`ChebyNet`, `LJacobiNet`, `S-JacobiNet`) perform poorly.
Even the adaptive `L-JacobiNet` fails to outperform its static counterparts, suggesting the $L_{hat} = L_{sym} - I$ mapping, centered at 0, is inherently biased towards low-pass responses and cannot be effectively "warped" to model high-frequency heterophilic signals.
\item The $[0, \infty)$ domain AOPFs (`MeixnerNet`, `LaguerreNet`, `KrawtchoukNet`) achieve SOTA results, with `MeixnerNet` being the clear winner.
\end{itemize}
This provides strong evidence that the $[0, \infty)$ domain (using $L_{scaled} = 0.5 L_{sym}$) is mathematically better suited for learning the band-pass filters required for heterophily, a finding consistent with recent GSP analysis.
\subsection{Trade-off 2: Stability vs. Spectral Domain}
We next test the stability of unbounded $O(k^2)$ filters (`LaguerreNet`, `L-JacobiNet`) at high polynomial degrees ($K$).
\begin{table}[htbp]
\caption{Test accuracies (\%) vs. $K$ (Over-smoothing) on PubMed (H=16).}
\label{tab:k_ablation}
\centering
\begin{tabular}{r c c c}
\toprule
$K$ & \textbf{ChebyNet} & \textbf{LJacobiNet} & \textbf{LaguerreNet} \\
\midrule
2 & 0.7780 & 0.7660 & 0.7640 \\
3 & 0.7640 & 0.7620 & 0.7660 \\
5 & 0.6480 & 0.7640 & \textbf{0.7980} \\
10 & 0.6260 & 0.7540 & 0.7860 \\
15 & 0.6010 & 0.7870 & 0.7500 \\
20 & 0.6770 & 0.7780 & 0.7560 \\
25 & 0.6140 & \textbf{0.7850} & \textbf{0.1800 (Collapse)} \\
30 & 0.6180 & 0.7800 & 0.1800 (Collapse) \\
\bottomrule
\end{tabular}
\end{table}

\begin{figure}[htbp]
\centerline{\includegraphics[width=\columnwidth]{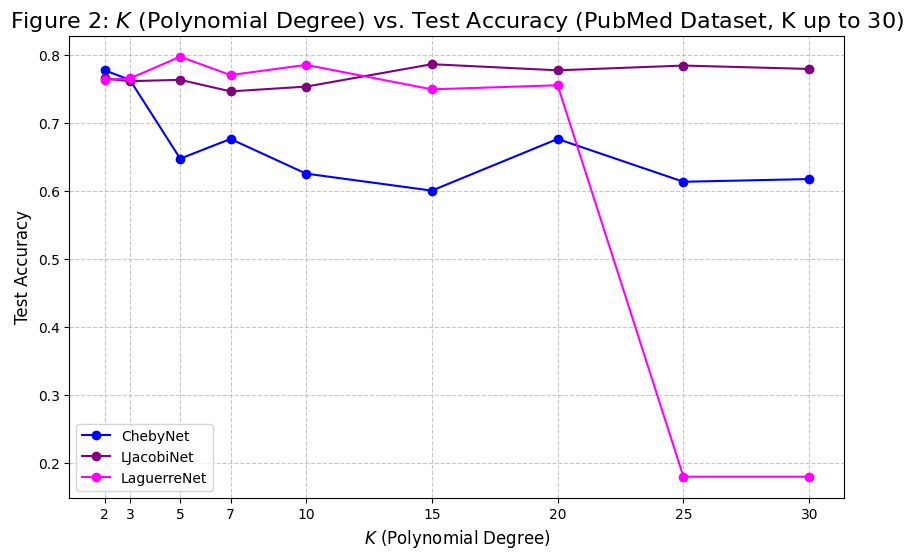}}
\caption{Figure 2: $K$ (Polynomial Degree) 
vs. Test Accuracy (PubMed). `ChebyNet` (blue) collapses at $K=5$. `LaguerreNet` (magenta) is stable until $K=25$, where it catastrophically collapses.
`L-JacobiNet` (purple) remains perfectly stable up to $K=30$.}
\label{fig:k_ablation}
\end{figure}

\textbf{Analysis of Stability Results:}
Table \ref{tab:k_ablation} and Figure \ref{fig:k_ablation} reveal the second critical trade-off.
\begin{itemize}
    \item `ChebyNet` (Static, Unstable) collapses immediately at $K=5$, demonstrating its inherent instability at high degrees.
    \item `LaguerreNet` (Adaptive, $O(k^2)$), operating on the semi-infinite $[0, \infty)$ domain, shows initial robustness due to `LayerNorm` but cannot maintain stability, catastrophically collapsing at $K=25$.
    \item `L-JacobiNet` (Adaptive, $O(k^2)$), operating on the finite $[-1, 1]$ domain, remains perfectly stable up to $K=30$, achieving SOTA results at $K=25$.
\end{itemize}
This suggests a key insight consistent with numerical analysis literature 
\cite{Hale2013, Rebollo2019}: for high-degree polynomials, the stability provided by a finite domain (Jacobi) is mathematically superior to that of a semi-infinite domain (Laguerre) \cite{Provost2009}.
`LayerNorm` alone is insufficient to stabilize unbounded $O(k^2)$ growth on an infinite domain, but it \textit{is} sufficient on a finite one.
\subsection{Trade-off 3: Adaptation vs. Stabilization in the $[-1, 1]$ Domain}
Finally, we analyze our core hypothesis: is `L-JacobiNet`'s adaptivity ($\alpha, \beta$) the key, or is it stabilization (`LayerNorm`)?
We compare `L-JacobiNet` (Adaptive+Stable) vs. `S-JacobiNet` (Static+Stable) vs. `ChebyNet` (Static+Unstable).
\begin{table}[htbp]
\caption{Test accuracies (\%): ChebyNet Genelleme Analizi (K=3).}
\label{tab:cheby_gen}
\centering
\begin{tabular}{l c c c}
\toprule
\textbf{Dataset} & \textbf{ChebyNet} & \textbf{S-JacobiNet} & \textbf{LJacobiNet} \\
 & (Static, Unstable) & (Static, Stable) & (Adaptive, Stable) \\
\midrule
Cora & 0.7730 & \textbf{0.7970} & 0.7870 \\
CiteSeer & \textbf{0.6730} & 0.6670 & 0.6300 \\
PubMed & 0.7510 & 0.7480 & \textbf{0.7840} \\
Texas & 0.7270 & \textbf{0.7946} & 0.7838 \\
Cornell & 0.6486 & \textbf{0.6622} & 0.6135 \\
\bottomrule
\end{tabular}
\end{table}

\begin{table}[htbp]
\caption{Öğrenilen Jacobi Parametreleri ($\alpha, \beta$) (K=3).}
\label{tab:learned_params}
\centering
\begin{tabular}{l c c}
\toprule
\textbf{Dataset} & \textbf{Learned $\alpha$} & \textbf{Learned $\beta$} \\
\midrule
Cora & -0.1641 & -0.4143 \\
CiteSeer & -0.2053 & 
-0.3814 \\
PubMed & -0.2618 & -0.3295 \\
Texas & -0.2705 & -0.3208 \\
Cornell & -0.2762 & -0.3143 \\
\bottomrule
\end{tabular}
\end{table}

\textbf{Analysis of Adaptation vs. Stabilization:}
This is our most significant finding, revealed in Table \ref{tab:cheby_gen}.
\begin{itemize}
    \item `S-JacobiNet` (Static+Stable) outperforms the standard, unstable `ChebyNet` baseline on 4/5 datasets.
This shows that `ChebyNet` is fundamentally limited by a lack of stabilization, a critical fact overlooked by prior work.
\item `S-JacobiNet` (Static+Stable) also outperforms our adaptive `L-JacobiNet` (Adaptive+Stable) on 4/5 datasets.
    \item Table \ref{tab:learned_params} confirms that `L-JacobiNet` \textit{is} learning;
its $\alpha, \beta$ parameters successfully deviate from the static $-0.5$ (Chebyshev) point.
\end{itemize}
This combination implies that for the $[-1, 1]$ domain, the benefit of stabilization (`LayerNorm`) is \textit{greater} than the benefit of adaptation ($\alpha, \beta$).
The adaptivity, while active (Table \ref{tab:learned_params}), appears to lead to overfitting on these datasets, making the simpler, stabilized-static `S-JacobiNet` a more robust model.
\section{Discussion and Future Work}
Our foundational analysis of the AOPF class, culminating in the `L-JacobiNet` and `S-JacobiNet` experiments, reveals critical trade-offs for GNN filter designers.
This analysis moves beyond finding a single "best" filter and provides a framework for selecting the correct polynomial basis and design paradigm.
\subsection{The Heterophily vs. Stability Trade-off}
Our results demonstrate a clear trade-off:
\begin{itemize}
    \item \textbf{For Heterophily:} The $[0, \infty)$ domain (e.g., `MeixnerNet`, `LaguerreNet`) is the superior choice.
The ability to learn a band-pass filter on this domain (consistent with \cite{HeroFilter2025, SplitGNN2023}) is more effective than any adaptation on the $[-1, 1]$ domain.
\item \textbf{For High-K Stability:} The $[-1, 1]$ domain (e.g., `L-JacobiNet`) is the superior choice.
Its finite, bounded nature provides a numerically stable foundation (consistent with \cite{Hale2013, Rebollo2019}) that allows `LayerNorm` to stabilize unbounded $O(k^2)$ coefficients up to $K=30$, whereas the semi-infinite domain of `LaguerreNet` eventually fails \cite{Provost2009}.
\end{itemize}

\subsection{The `S-JacobiNet` Discovery (Adaptation vs. Stabilization)}
Our most significant finding is the surprising power of `S-JacobiNet` (Table \ref{tab:cheby_gen}).
The GNN community has largely assumed `ChebyNet`'s flaw was its \textit{static basis}.
Our results show `ChebyNet`'s flaw was its \textit{lack of stabilization}.
`S-JacobiNet` (functionally `ChebyNet` + `LayerNorm`) outperforms its adaptive counterpart `L-JacobiNet` on 4/5 datasets.
We identify `S-JacobiNet` as a powerful, simple, and overlooked baseline that should be adopted in future GNN research.
\subsection{The Bias-Variance Trade-off: Why S-JacobiNet Excels}
The superiority of `S-JacobiNet` over `L-JacobiNet` can be explained by the classic bias-variance trade-off.
\begin{itemize}
    \item \textbf{`L-JacobiNet` (High Variance):} By making $\alpha$ and $\beta$ learnable, `L-JacobiNet` gains immense flexibility to change the entire polynomial basis.
On small benchmark datasets (Cora, CiteSeer, etc.), this high capacity (low bias) allows the model to overfit to the training data, as seen in its lower test accuracy.
\item \textbf{`S-JacobiNet` (Low Variance):} `S-JacobiNet` fixes $\alpha=\beta=-0.5$, effectively locking the model into the Chebyshev basis.
This acts as a \textit{strong regularizer}, significantly reducing the model's variance (i.e., increasing its bias).
\end{itemize}
Our results suggest that the Chebyshev basis (high bias) is "\textit{good enough}" for these tasks, and the primary bottleneck was never the basis, but rather the numerical instability (high variance) of the filter, which `LayerNorm` solves.
This "simplicity-first" finding aligns with the success of other low-variance models like `APPNP` \cite{gasteiger2019predict} and `GPR-GNN` \cite{chien2021adaptive}, which also leverage simple, static propagation schemes.
\subsection{Future Work}
This trade-off suggests a "Dual-Domain" or "Split-Spectrum" GNN as a logical next step.
Such a model could use a `LaguerreNet` or `MeixnerNet` branch to process high-frequency (heterophilic) signals and a separate `S-JacobiNet` (not `L-JacobiNet`) branch to process low-frequency (homophilic) signals deeply and stably.
A key research question would be how to route the signals;
this could be achieved with a learnable gating mechanism or a spectral attention mechanism, similar to `GAT` \cite{velickovic2018graph}, that learns to assign different frequency components of the input signal to the appropriate filter branch.
\section{CONCLUSION}
We introduced `L-JacobiNet`, the adaptive generalization of `ChebyNet`, and `S-JacobiNet`, its stabilized-static counterpart, to analyze the AOPF class.
Our experiments did not reveal a single "best" filter. Instead, we discovered a fundamental \textbf{domain-specific trade-off} in GNN filter design.
We conclude that the $[0, \infty)$ domain (`LaguerreNet`/`MeixnerNet`) is the correct choice for \textbf{heterophily}, while the $[-1, 1]$ domain (`L-JacobiNet`) is the correct choice for \textbf{high-K stability}.
Finally, we showed that `ChebyNet`'s primary flaw is not its static basis but its lack of stabilization.
Our static `S-JacobiNet` model (ChebyNet + LayerNorm) emerged as an overlooked and highly competitive SOTA baseline, suggesting that stabilization is more critical than adaptation in the $[-1, 1]$ domain.


\begin{thebibliography}{99}
\itemsep 1pt

\bibitem{shuman2013emerging}
D. I. Shuman, S. K. Narang, P. Frossard, A. Ortega, and P. Vandergheynst, "The emerging field of signal processing on graphs," \textit{IEEE Signal Processing Magazine}, vol.
30, no. 3, pp. 83-98, 2013.

\bibitem{defferrard2016convolutional}
M. Defferrard, X. Bresson, and P. Vandergheynst, "Convolutional neural networks on graphs with fast localized spectral filtering," in \textit{Advances in Neural Information Processing Systems (NIPS)}, 2016.

\bibitem{kipf2017semi}
T. N. Kipf and M. Welling, "Semi-supervised classification with graph convolutional networks," in \textit{Intl.
Conf. on Learning Representations (ICLR)}, 2017.

\bibitem{koekoek2010hypergeometric}
R. Koekoek, P. A. Lesky, and R. F. Swarttouw, \textit{Hypergeometric Orthogonal Polynomials and Their q-Analogues}.
Springer, 2010.

\bibitem{ba2016layer}
J. L. Ba, J. R. Kiros, and G. E. Hinton, "Layer normalization," \textit{arXiv preprint arXiv:1607.06450}, 2016.

\bibitem{goksu2025meixnernet}
H. Göksu, "MeixnerNet: Adaptive and robust spectral graph neural networks with discrete orthogonal polynomials," \textit{IEEE Signal Processing Letters}, 2025. (Submitted for review).
\bibitem{goksu2025krawtchouknet}
H. Göksu, "KrawtchoukNet: A unified GNN solution for heterophily and over-smoothing with adaptive bounded polynomials," \textit{IEEE Transactions on Neural Networks and Learning Systems}, 2025. (Submitted for review).
\bibitem{goksu2025laguerrnet}
H. Göksu, "LaguerreNet: Advancing a Unified Solution for Heterophily and Over-smoothing with Adaptive Continuous Polynomials," \textit{IEEE Transactions on Signal Processing}, 2025. (Submitted for review).
\bibitem{zhu2020beyond}
J. Zhu, Y. Wang, H. Wang, J. Zhu, and J. Tang, "Beyond homophily in graph neural networks: Current limitations and open challenges," in \textit{Proc.
ACM SIGKDD Intl. Conf. on Knowledge Discovery \& Data Mining (KDD)}, 2020.

\bibitem{li2018deeper}
Q. Li, Z. Han, and X. Wu, "Deeper insights into graph convolutional networks for semi-supervised learning," in \textit{AAAI Conf.
on Artificial Intelligence}, 2018.

\bibitem{pei2020geomgcn}
H. Pei, B. Wei, K. C. C. Chang, Y. Lei, and B. Yang, "Geom-GCN: Geometric graph convolutional networks," in \textit{Intl.
Conf. on Learning Representations (ICLR)}, 2020.

\bibitem{velickovic2018graph}
P. Veličković, G. Cucurull, A. Casanova, A. Romero, P. Liò, and Y. Bengio, "Graph attention networks (GAT)," in \textit{Intl.
Conf. on Learning Representations (ICLR)}, 2018.

\bibitem{gasteiger2019predict}
J. Gasteiger, A. Bojchevski, and S. Günnemann, "Predict then propagate: Graph neural networks meet personalized pagerank," in \textit{Intl.
Conf. on Learning Representations (ICLR)}, 2019.

\bibitem{chen2020simple}
M. Chen, Z. Wei, Z. Huang, B. Ding, and Y. Li, "Simple and deep graph convolutional networks," in \textit{Intl.
Conf. on Machine Learning (ICML)}, 2020.

\bibitem{xu2018representation}
K. Xu, C. Li, Y. Tian, T. Sonobe, K. Kawarabayashi, and S. Jegelka, "Representation learning on graphs with jumping knowledge networks," in \textit{Intl.
Conf. on Machine Learning (ICML)}, 2018.

\bibitem{levie2018cayleynets}
R. Levie, F. Monti, X. Bresson, and M. M. Bronstein, "CayleyNets: Graph convolutional neural networks with complex rational spectral filters," \textit{IEEE Transactions on Signal Processing}, vol.
67, no. 1, pp. 97-112, 2018.

\bibitem{sen2008collective}
P. Sen, G. Namata, M. Bilgic, L. Getoor, B. Galligher, and T. Eliassi-Rad, "Collective classification in network data," \textit{AI Magazine}, vol.
29, no. 3, p. 93, 2008.

\bibitem{HeroFilter2025}
G. Li, J. Yang, and S. Liang, "HeroFilter: Adaptive Spectral Graph Filter for Varying Heterophilic Relations," \textit{arXiv preprint arXiv:2510.10864}, 2S025.
\bibitem{SplitGNN2023}
Y. Wang, Z. Wu, J. Xu, Z. Wang, and S. Zhang, "SplitGNN: Spectral Graph Neural Network for Fraud Detection against Heterophily," in \textit{Proc.
32nd ACM Intl. Conf. on Information and Knowledge Management (CIKM)}, 2023.

\bibitem{JacobiConv2022}
X. Wang and M. Zhang, "How Powerful are Spectral Graph Neural Networks," in \textit{Intl.
Conf. on Machine Learning (ICML)}, 2022.

\bibitem{ARMAConv2021}
F. M. Bianchi, D. Grattarola, C. Alippi, and L. Livi, "Graph neural networks with convolutional ARMA filters," \textit{IEEE Transactions on Pattern Analysis and Machine Intelligence (TPAMI)}, vol.
45, no. 5, pp. 5999-6011, 2021.

\bibitem{chien2021adaptive}
E. Chien, J. Liao, W. H. Chang, and C. K. Yang, "Adaptive graph convolutional neural networks (GPR-GNN)," in \textit{Intl.
Conf. on Learning Representations (ICLR)}, 2021.
\label{GPRGNN2021} 

\bibitem{ResChebyNet2024}
Y. Liu, S. Gao, Z. Xia, F. Li, and S. Cui, "Efficient Multi-View Graph Convolutional Network with Self-Attention for Multi-Class Motor Imagery Decoding," \textit{IEEE Trans.
Neural Syst. Rehabil. Eng.}, vol. 32, pp. 2486-2495, 2024.

\bibitem{LSGAT2022}
B. Liu, Y. Wang, K. Li, P. Li, G. Zhang, and Z. Zhang, "Simple and Deep Graph Attention Networks," in \textit{Intl.
Conf. on Learning Representations (ICLR)}, 2022.

\bibitem{Hale2013}
N. Hale and A. Townsend, "Fast and Accurate Computation of Gauss-Legendre and Gauss-Jacobi Quadrature Nodes and Weights," \textit{SIAM Journal on Scientific Computing}, vol.
35, no. 2, pp. A652-A674, 2013.

\bibitem{Rebollo2019}
M. Rebollo, A. B. J. Kuijlaars, and A. Martínez-Finkelshtein, "Noniterative Computation of Gauss-Jacobi Quadrature," \textit{SIAM Journal on ScientificComputing}, vol.
41, no. 2, pp. A1269-A1291, 2019.

\bibitem{Provost2009}
S. B. Provost and H. Sabzalian, "Moment-Based Density Approximants," \textit{Communications in Statistics - Simulation and Computation}, vol.
38, no. 6, pp. 1157-1173, 2009.

\bibitem{he2021bernnet}
  M. He, Z. Wei, Zhewei and H. Xu and others
  "Bernnet: Learning arbitrary graph spectral filters via bernstein approximation"
\textit{Advances in neural information processing systems},
vol.
34,
pp. 14239--14251,
2021


\end{thebibliography}
\end{document}